\documentclass{article}
\usepackage{spconf,amsmath,graphicx}

\def\@fnsymbol#1{\ensuremath{\ifcase#1\or \dagger\or \ddagger\or
   \mathsection\or \mathparagraph\or \|\or **\or \dagger\dagger
   \or \ddagger\ddagger \else\@ctrerr\fi}}
    \makeatother

\title{SPADE: Self-supervised Pretraining for Acoustic DisEntanglement}
\name{John Harvill$^{1*}$, Jarred Barber$^{2\dagger}$, Arun Nair$^{2}$, Ramin Pishehvar$^{2}$\thanks{$^*$Work done as part of an internship at Amazon Alexa AI. ~ ~ ~ ~ ~ ~ ~ ~ ~ ~ ~ ~ ~ ~ ~ ~ ~ ~ ~ ~   $^\dagger$ Now at Google Research}}
\address{$^1$University of Illinois Urbana-Champaign, USA\\
$^2$Amazon Alexa AI, USA}
\begin{document}
\ninept
\maketitle
\begin{abstract}


Self-supervised representation learning approaches have grown in popularity due to the ability to train models on large amounts of unlabeled data and have demonstrated success in diverse fields such as natural language processing, computer vision, and speech. Previous self-supervised work in the speech domain has disentangled multiple attributes of speech such as linguistic content, speaker identity, and rhythm. In this work, we introduce a self-supervised approach to disentangle room acoustics from speech and use the acoustic representation on the downstream task of device arbitration. Our results demonstrate that our proposed approach significantly improves performance over a baseline when labeled training data is scarce, indicating that our pretraining scheme learns to encode room acoustic information while remaining invariant to other attributes of the speech signal. 

\end{abstract}
\section{Introduction}
\label{sec:intro}


Disentanglement of speech into its multiple components is a fundamental problem in signal processing with applications in voice conversion \cite{qian2019autovc, qian2020unsupervised, wang2021adversarially, lian2022robust}, automatic speech recognition \cite{qian2022contentvec, ebbers2021contrastive, hsu2018unsupervised, khurana2019factorial}, speaker recognition \cite{nagrani2020disentangled}, and privacy preservation in speech \cite{aloufi2020privacy}. The goal of disentanglement is to separate different attributes of the speech signal such that the final signal representation will be invariant to attributes not relevant to the downstream task. Previous work has focused on invariance towards acoustic content\footnote{In this paper, “acoustic content", or “acoustics" refers to the information in an audio signal related to the Room Impulse Response (RIR) at a particular location in a room when the source audio is played from a different, fixed location.} so that target attributes like speaker identity or linguistic content will be emphasized. In this paper, we specifically want to preserve acoustic content and extract representations that are invariant to all other attributes in a self-supervised fashion.
These representations are then used for the task of device arbitration \cite{barber2022end}.

The device arbitration task has arisen recently due to the ubiquity of smart voice assistant-enabled devices, which we refer to as “voice assistants" (VA) or “devices" interchangeably. Many households now have multiple VAs in the same room, leading to ambiguity with respect to which device should interact with the user. When the user wishes to begin interaction with a VA, they must first utter a wakeword (“Alexa", “Hey Google", etc.) which is then recorded by all devices in the room. For the most natural user experience, only the intended device from the user should wake up and continue to interact with the user. This leads to the device arbitration problem: \textit{given $N$ recordings of a source audio, where each recording comes from a VA, determine which VA is the intended one}.

Note that device arbitration is closely related to the well-studied source localization problem \cite{chen2002source}. 
Time Difference Of Arrival (TDOA) \cite{gustafsson2003positioning} is an effective technique that solves source localization for audio signals but unfortunately requires large arrays of microphones not present on VAs. There are several other constraints imposed by modern VAs that make device arbitration a non-trivial problem: (1) The positions of each VA are unknown and could change over time (moving the device from the living room to the kitchen). (2) The acoustic environment of the VAs is unknown. (3) Clock synchronization between devices is not always available, making TDOA between devices infeasible. Given these constraints, the device arbitration task must be solved by only relying on the room acoustic information present in each audio recording.

Motivated by these constraints, we propose SPADE: Self-supervised Pretraining for Acoustic DisEntanglement. SPADE is a pretraining technique that disentangles acoustic information from speech signals by using multiple views of a source audio without the need for labels. We find that when used in combination with previous work on device arbitration \cite{barber2022end}, SPADE leads to improved performance when less labeled training data is present. Given that labeled data is difficult to collect for this task, SPADE is an invaluable technique for improving performance on device arbitration at zero additional inference 
cost.

 The remainder of our paper is organized as follows: In Section 2 we discuss prior work on device arbitration and related tasks. In Section 3, we discuss the data used in our experiments and the simulation process from which it is generated. In Section 4, we detail our proposed pretraining approach. In Section 5, we discuss our experimental setup and in Section 6 we discuss results. The paper ends with conclusions and suggestions for future work in Section 7.


\section{Prior Work}
\label{sec:prior_work}

While device arbitration is a relatively new problem, it is closely related to source localization \cite{chen2002source, gustafsson2003positioning}. The goal of source localization is to determine the position of the object emitting sound, i.e. the source. Current techniques rely on large arrays of microphones which are not available given our problem setup. Previous work \cite{barber2022end} also demonstrated that directly predicting source distances from each device and making arbitration decisions based on those distances resulted in worse performance. 

The goal in arbitration is to select an attended device the user is speaking to based on distance and direction. Common techniques compare signal attributes like Signal-to-Noise Ratio (SNR), estimated distance between source and microphone, cross-correlation, etc. \cite{kumatani2011channel}. The goal of device arbitration is different from channel selection \cite{kumatani2011channel, wolf2014channel}  since the optimal device is simply defined as that which is closest to the user or as the attended device (defined as the one in the look direction of the user or any other acoustical or visual relevant cue), and not that which leads to better performance on another downstream task. The differences in problem setup and objective between device arbitration and source localization show that device arbitration should be studied separately and has its own set of unique challenges that make it an interesting problem.

We used as a baseline, the machine learning-based device arbitration proposed by Barber et al. \cite{barber2022end}.
The authors proposed an end-to-end approach to train a neural network to perform device arbitration. Their model consists of a small convolutional feature extractor that runs locally on each device, and a larger arbitration network that runs in the cloud to make the final decision. Results demonstrated significantly improved performance across a variety of room conditions over a simple energy-based approach. 




\section{Dataset Simulation}
\label{sec:dataset}


Currently, there is no large-scale dataset for device arbitration with known ground truth labels, so we run our experiments with simulated data following the three main steps described in Barber et al. \cite{barber2022end}: 


\begin{itemize}
    \item \textbf{Sample scenario:} Sample a room from a variety of acoustic settings (room length/width, noise sources) as well as device/speaker positions within the room. Given device/speaker positions, generate the arbitration label based on smallest Euclidean distance from device to speaker.
    \item \textbf{Generate RIR:} Given the sampled scenario, generate a Room Impulse Response (RIR) for each device in the scenario using an acoustic simulator.
    \item \textbf{Generate audio:} Convolve speech utterances with generated RIRs and mix with noise for each device. After this step, we have an artificial dataset of device arbitration audio and corresponding ground truth labels.
\end{itemize}
We use the Image Source Method (ISM) \cite{dance1997complete} for data simulation and source audio from \cite{barber2022end}, but have updated the scene sampling hyperparameters to those in Table \ref{tab:hparams_sampling}.

\begin{table}[]
    \centering
    \begin{tabular}{|c|c|} \hline
    Parameter & Distribution \\ \hline
    Room length/width (m) & Uniform(3.0, 10.0) \\ \hline
    Room height (m) & Uniform(2.5, 6.0) \\ \hline
    Reverberation time (s) & Beta(1.1, 3.0) \\ \hline
    Number of devices & ShiftedPoisson(m=3,l=2,h=15) \\ \hline
    Number of noise sources & ShiftedPoisson(m=2,l=1,h=5) \\ \hline
    Speech level (dB SPL) & Uniform(55.0, 70.0) \\ \hline
    Noise level (dB SPL) & Uniform(50.0, 70.0) \\ \hline
    \end{tabular}
    \caption{Hyperparameters for scene sampling. For the ShiftedPoisson distribution we denote mean with “m", low with “l" and high with “h." 
    }
    \label{tab:hparams_sampling}
\end{table}



\section{Method}
\label{sec:method}

Our arbitration model is based on that proposed by Barber et al. \cite{barber2022end} and is composed of two components: the feature encoder and the classifier that makes the arbitration decision. Prior to end-to-end training of the encoder and classifier, 
we pretrain the encoder using two schemes: contrastive and reconstructive pretraining. Our preprocessing pipeline and pretraining schemes are discussed in the following subsections.


\subsection{Audio Preprocessing}

Audio is first transformed to log-filterbank energy (LFBE) features, where the spectrogram is computed using a 25ms frame size and a frame skip of 10ms. The Mel transform is applied to the spectrogram with 64 Mel bands, followed by the log transform. LFBE features are 
mean and variance normalized before downstream processing by neural models. 

\subsection{Encoder Architecture}\label{sec:encoder_arch}

\begin{figure}
    \centering
    \includegraphics[width=8.3cm, height=6.6cm]{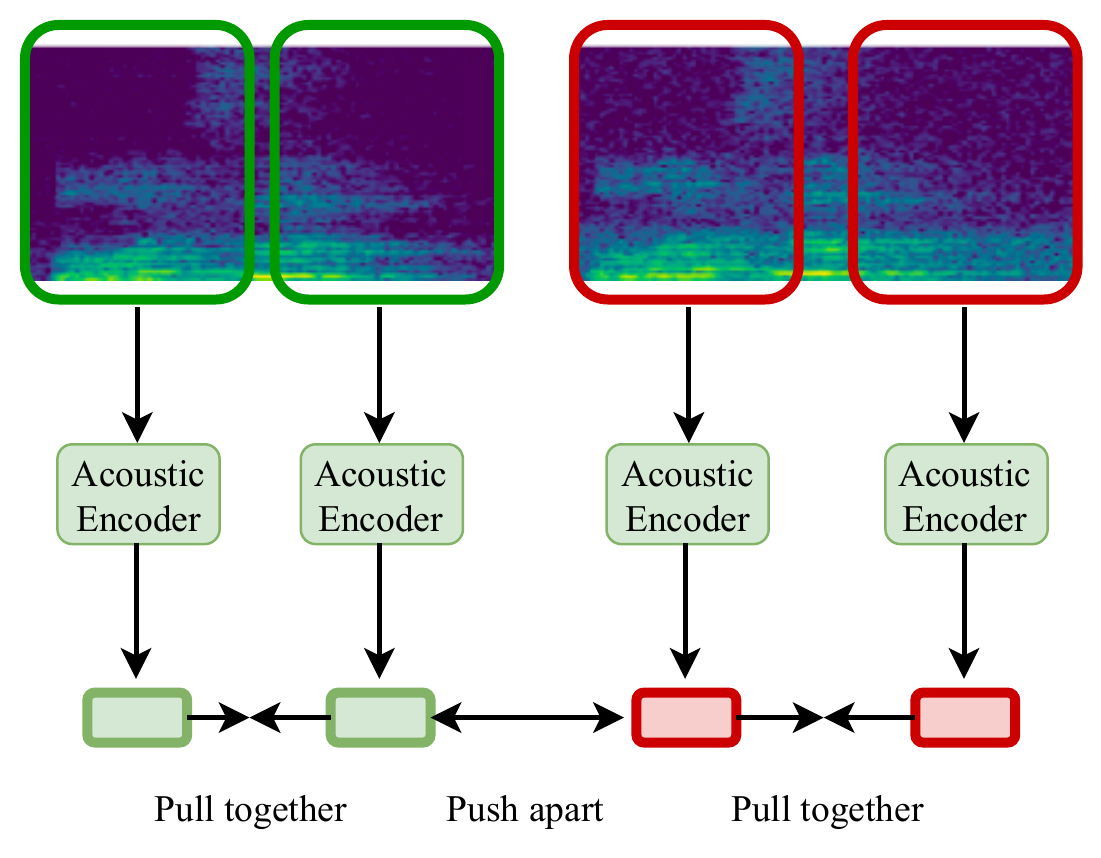}
    \caption{Contrastive pretraining scheme: Acoustic encoder has shared weights.}
    \label{fig:contrastive}
\end{figure}

The encoder architecture is a residual convolutional model composed of 18 convolutional layers, with batch normalization and the ReLU activation function applied to each layer.
The encoder model is the same for all approaches discussed in this paper (pretraining, baseline).
This network produces a sequential feature representation of the input with much smaller temporal resolution than the input LFBE features. During pretraining we use a small Transformer \cite{vaswani2017attention} network to map the sequence of vectors output by the encoder to a single vector. This network is discarded after pretraining but is implicit as part of the encoder in the following pretraining discussions and Figs. \ref{fig:contrastive} and \ref{fig:recon}.


\subsection{Contrastive Pretraining}

Contrastive loss functions have been shown to create high-quality representations across a variety of domains like speech and natural language processing \cite{baevski2020wav2vec, gao-etal-2021-simcse}. This family of loss functions operates by assuming data are similar or dissimilar with respect to a particular attribute and encouraging embeddings of the data to reflect these relationships via distance in an embedding space. In a device arbitration scene, audio from the same device has the same room acoustic properties, while audio from different devices will have different acoustic properties.\footnote{ This is due to the assumption that the devices are in different locations.} Since we want to encode room acoustic information, we can encourage embeddings of audio from the same device to be similar and simultaneously encourage embeddings of audio from different devices to be orthogonal (see Fig. \ref{fig:contrastive}). Since speech content will be the same across all $N$ recordings from a given room, the embedding should be invariant to speech content and only encode acoustic information.

Let us denote the audio recorded by device $a$ as $x^a$. For this pretraining approach, we create a fixed-length embedding $z^a$ of $x^a$ by passing through the encoder\footnote{The Transformer network discussed in Section \ref{sec:encoder_arch} is implied here to follow the encoder to create the fixed-length representation.}, i.e. $z^a = \text{encoder}(\text{LFBE}(x^a))$. Then $z^a$ is normalized to have magnitude one. We can denote a continuous slice of $x^a$ from time $t_1$ to $t_2$ as $x^a_{t_1:t_2}$ where $t_2 > t_1$ and its corresponding fixed-length embedding as $z^a_{t_1:t_2}$. Given $N$ recordings $x^1,x^2,...,x^N$ that are zero-padded to the same length $T$ and a splitting index $t_{split}$ such that $0<\frac{T}{2}-\epsilon<t_{split}<\frac{T}{2}+\epsilon<T$, where $\epsilon$ is a small random jitter, we arrive at our loss function $\mathcal{L}_{C}$ for one arbitration scenario:

\begin{equation}
    \mathcal{L}_1 = \sum_{i=1}^N \sum_{j=1}^N | \langle z^i_{0:t_{split}} , z^j_{t_{split}:T} \rangle - \delta_{ij}|
\end{equation}

\begin{equation}
    \mathcal{L}_2 = \sum_{i=1}^N \sum_{j\neq i} | \langle z^i_{0:t_{split}} , z^j_{0:t_{split}} \rangle | + | \langle z^i_{t_{split}:T} , z^j_{t_{split}:T} \rangle |
\end{equation}

\begin{equation}
    \mathcal{L}_{C} = \mathcal{L}_1 + \mathcal{L}_2
\end{equation}
where $\delta_{ij}$ is the Kronecker delta function. Note that $\mathcal{L}_1$ encourages the two halves of the same audio recording to map to the same embedding. The $\mathcal{L}_2$ term provides stronger supervision for invariance to speech content by encouraging different recordings of the same respective halves of the audio to be orthogonal.

\subsection{Reconstructive Pretraining}

Disentanglement of different attributes in speech has been accomplished previously using autoencoding with an information bottleneck \cite{qian2019autovc, qian2020unsupervised}. We take a similar approach to disentangle room acoustic information from speech in a self-supervised fashion, making the assumption that the room acoustic properties are constant\footnote{This stationarity assumption may not be true in all cases (people/pets moving around), but it is reasonable given that audio is only recorded over a two-second interval.} for the duration of the wakeword audio ($\sim$2s).

\begin{figure}[t!]
    \centering
    \includegraphics[width=8.3cm, height=10.0cm]{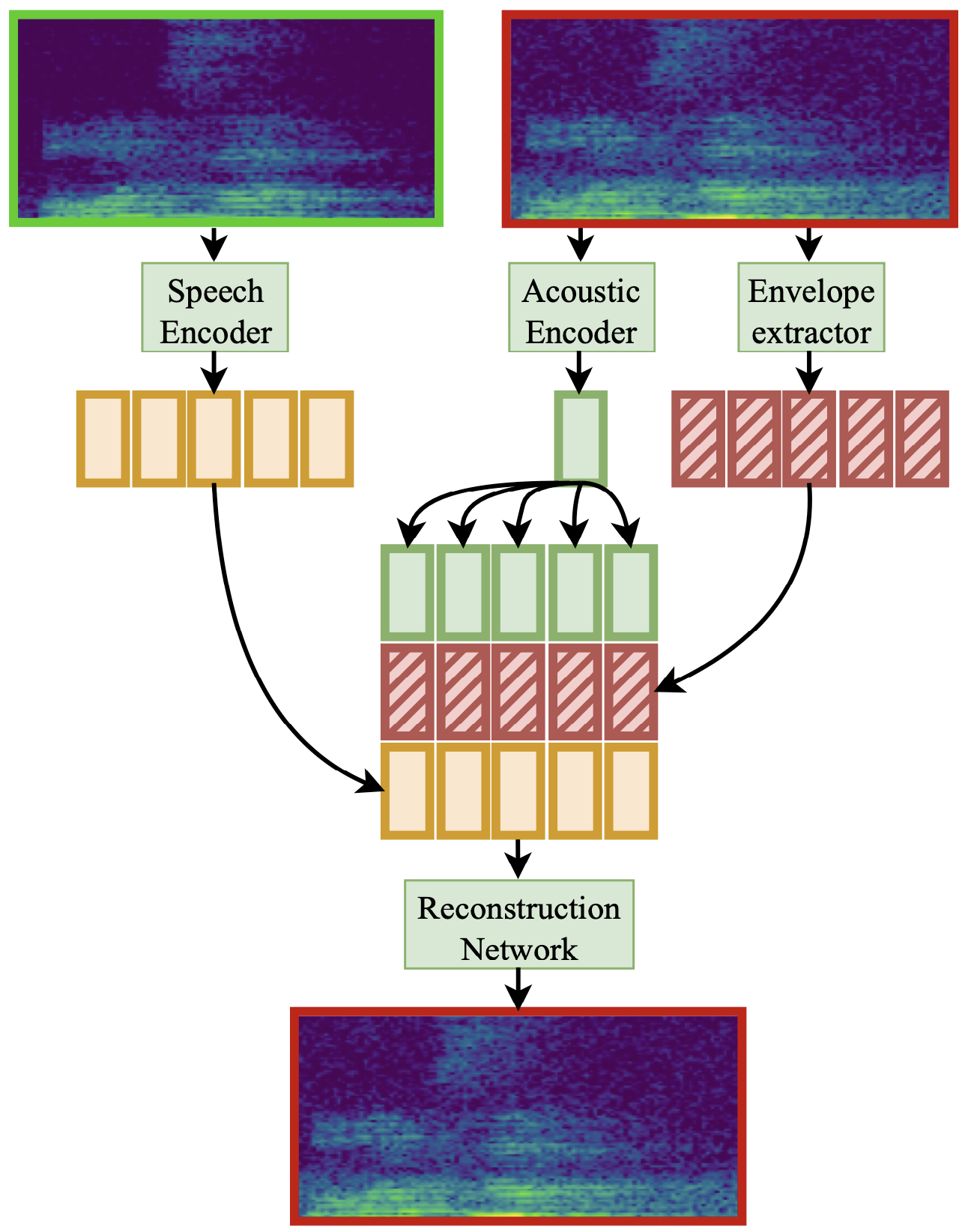}
    \caption{Reconstruction pretraining scheme: Speech encoder, envelope, and reconstruction network have same time axis dimension as the LFBE time axis. Acoustic encoder creates fixed-length embedding that is copied along time axis of speech encoder and envelope output. Diagonally-shaded units are from non-learnable modules.}
    \label{fig:recon}
\end{figure}

Our model consists of a speech encoder, an acoustic encoder, and a reconstruction decoder. The speech encoder $S(\cdot)$ and reconstruction decoder $R(\cdot)$ are Transformers \cite{vaswani2017attention} and each produce a sequence of vectors. We design the acoustic encoder $A(\cdot)$ to produce a fixed-dimensional embedding due to the stationarity assumption of the acoustics and the need to create an information bottleneck.
Given that $N$ recordings of a source audio all contain identical speech content, the speech representation $s_i=S(x^i)$ should be the same for each recording, i.e. $s^1\approx s^2 \approx...\approx s^N$. To encourage this, we reconstruct one audio recording's LFBE features using its acoustic embedding and the speech embedding from another audio recording in the room. The information bottleneck enforced by creating a small fixed-dimensional embedding encourages the acoustic encoder only to represent information not common to all signals, i.e. room acoustics. Given that the $N$ recordings are not time-aligned, we provide alignment information to the decoder by extracting the envelope $env(x^i)$ of the LFBE features for $x^i$, which is the mean of the feature vector at each timestep:
\begin{equation}
    env(x^i)_k = \frac{1}{N} \sum_{q=1}^{N} \text{LFBE}(x^i)_{kq}
\end{equation}
where $k$ denotes the time axis, $q$ denotes the feature axis of the LFBE feature matrix, and $N=64$. Note that the envelope extractor is not a learnable module but rather a simple mean operation at each timestep to produce a low-dimensional representation of the LFBE features (feature vector $\rightarrow$ scalar). We can formally write our loss function $\mathcal{L}_R$ for one arbitration scenario as:




\begin{equation}
    \mathcal{L}_R = \frac{1}{N}\sum_{i=1}^N ||\text{LFBE}(x^i) - R(S(x^{\neq i}), A(x^i), env(x^i))||_2^2
\end{equation}
where $x^{\neq i}$ denotes a randomly-selected audio recording from $\{x^1, x^2, ..., x^N\}$ other than $x^i$. We form the input to $R(\cdot)$ by copying $A(x^i)$ along the time axis and concatenating it to $S(x^{\neq i})$ and $env(x^i)$. See Fig. \ref{fig:recon} for a diagram detailing this process.

\subsection{Arbitration Classifier Architecture}

The arbitration classifier was implemented previously \cite{barber2022end} as a Multilayer Perceptron (MLP) network, but we found further improvement using a self-attention network like the Transformer \cite{vaswani2017attention}.
For each device $i$, the encoder outputs a sequence of hidden states $h^i_1, h^i_2, ..., h^i_K$. For $N$ devices, the hidden states are concatenated along the time axis to form the sequence:

\begin{equation}
    H = h^1_1, h^1_2,...,h^1_K, h^2_1,h^2_2,...h^2_K,...,h^N_1,h^N_2,...h^N_K
\end{equation}
which is then passed through a network of self-attention layers\footnote{Positional encodings are added to the input.} to produce the sequence:

\begin{equation}
    G = g^1_1, g^1_2,...,g^1_K, g^2_1,g^2_2,...g^2_K,...,g^N_1,g^N_2,...g^N_K
\end{equation}
Each sequence $g^i_1, g^i_2, ..., g^i_K$ is then passed through a second network of self-attention to create a summary $G_i$ over time. Each $G_i$ is then passed through a two-layer feedforward neural network, outputting a scalar logit for device $i$. The logits are then passed through a softmax layer to produce the arbitration probabilities. The entire classification network is optimized using the crossentropy loss between the arbitration probabilities and the ground truth label distribution.



\subsection{Baseline}

The contributions of this paper are the self-supervised pretraining approaches, so as a baseline, we use the encoder and classifier networks discussed previously but do not pretrain the encoder. 

\section{Experiments}
\label{sec:experiments}

The training dataset consists of 300k arbitration scenarios. To demonstrate the effectiveness of pretraining for learning representations useful for device arbitration, we create datasets of exponentially decreasing size. For dataset $i$ the training set size $s_i=\lfloor S/4^i \rfloor$, where $S=$300k (full training set size). We choose $i\in\{0, 1, 2, 3\}$ such that the smallest training set consists of $\sim$ 4.7k scenarios.

\subsection{Experimental Procedure}

For each experiment, we choose the final arbitration model based on the checkpoint with the lowest validation loss and evaluate on a held-out test set. We have four experiment setups:

\begin{itemize}
    \item \textbf{Baseline:} Train encoder-classifier model end-to-end on each of the training data subsets of size $s_i$ for $i\in\{0, 1, 2, 3\}$.
    \item \textbf{Contrastive:} Pretrain (acoustic) encoder using contrastive approach on all available training data (300k scenarios, no labels involved). Pick best validation checkpoint as initialization for encoder and then finetune encoder-classifier model end-to-end on each training data subset of size $s_i$ for $i\in\{0, 1, 2, 3\}$.
    \item \textbf{Reconstructive:} Same as contrastive setup except that we use reconstructive pretraining.
    \item \textbf{Combo:} Pretrain using both contrastive and reconstructive pretraining. Loss function becomes $\mathcal{L} = \lambda\mathcal{L}_R + (1-\lambda)\mathcal{L}_C$ where we set $\lambda=0.5$. Finetune encoder-classifier model as in previous setups.
    
\end{itemize}



\section{Results}
\label{sec:results}

\begin{figure}
    \centering
    \includegraphics[width=7.0cm, height=5.4cm]{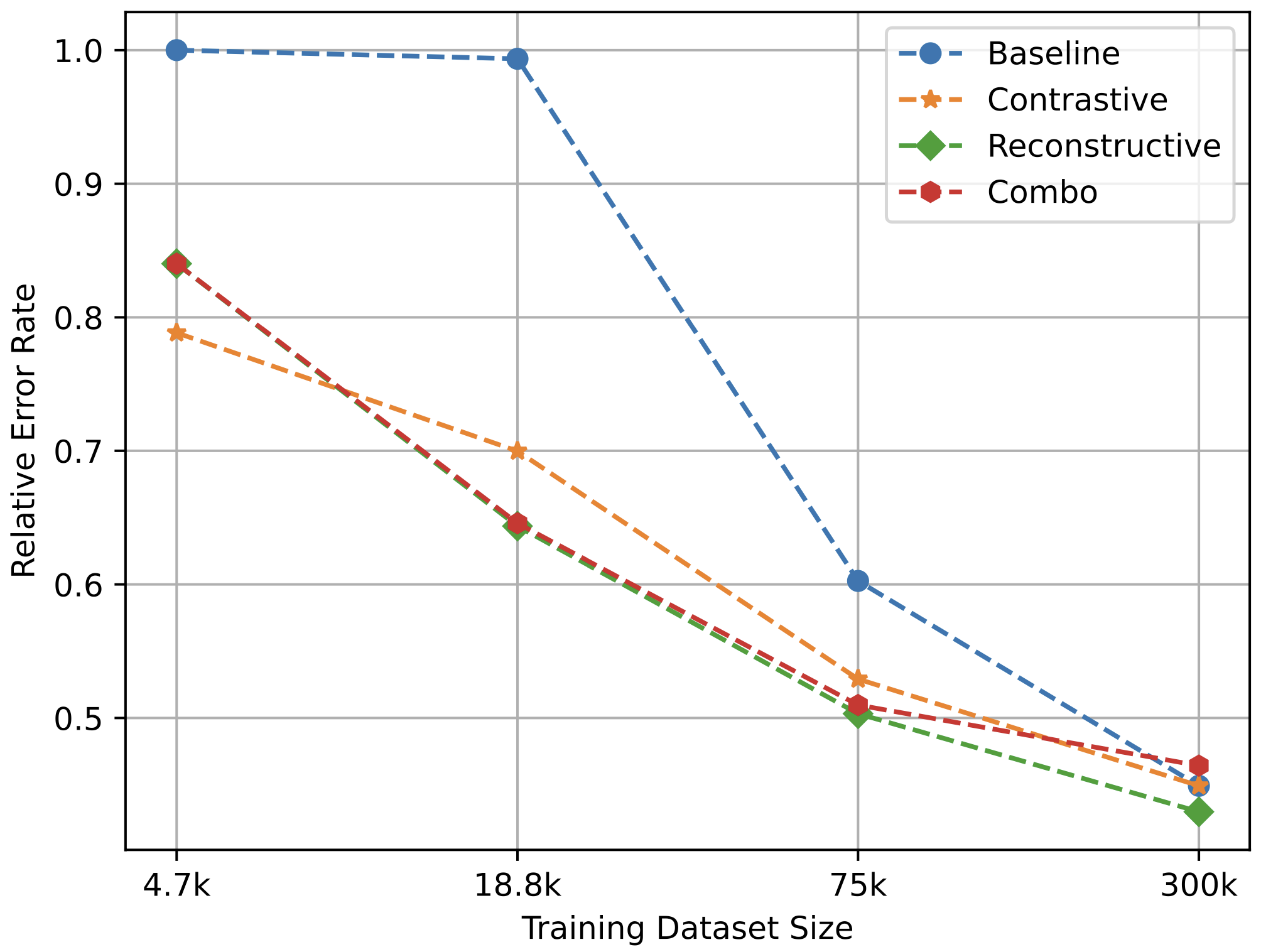}
    \caption{Relative Error Rate with respect to worst case.}
    \label{fig:wrt_worst_case}
\end{figure}

Results are presented here as \textit{relative error rate} with respect to the performance of the 4.7k baseline setting. Denoting the accuracy of the target method $m$ as $\text{acc}_{m}$ and the accuracy of the 4.7k baseline setting as $\text{acc}_{\text{base}}$, we compute relative error rate as:

\begin{equation}
    \text{err}_{\text{rel}} = \frac{1 - \text{acc}_{m}}{1 - \text{acc}_{\text{base}}}
\end{equation}
Our results are presented in Figure \ref{fig:wrt_worst_case}. The most important trend is that the pretraining approaches outperform the baseline by a larger margin when the training dataset is small. This demonstrates that our proposed pretraining schemes 
preserve acoustic information, learning features relevant to the device arbitration problem and are most beneficial when labeled training data is scarce. Initialization of the encoder with a pretrained checkpoint helps combat overfitting during the device arbitration training process.
We also find that the combination of contrastive and reconstructive pretraining does not lead to any noticeable improvement over either approach in isolation, indicating that both approaches may encode similar information.



\section{Conclusion}
\label{sec:conclusion}

In this paper we propose constrastive and reconstructive pretraining, two forms of self-supervised representation learning, that disentangle acoustic content from speech. Unlike previous work that has aimed to create representations that are invariant to acoustic content, we aim to encode acoustic content only 
and demonstrate its usefulness through the device arbitration problem. 
We find that both of our proposed pretraining approaches lead to improvement over the baseline, and that improvement is more significant when the labeled training dataset is small. This provides empirical evidence that our pretraining objectives lead to representations of acoustic content that can be useful for the device arbitration task even in the absence of a large training corpus.

Given that self-supervised techniques require no human annotations, it may be possible to apply our proposed approaches to other research problems. For example, our disentangled acoustic representations may be used for other acoustic tasks like room acoustic property estimation or acoustic adaptation for home theater. While not studied in this paper, the speech content representations learned from reconstructive pretraining may be valuable for acoustic-invariant applications like speaker recognition or ASR since they are designed to encode all information except acoustics.

\vfill\pagebreak

\bibliographystyle{IEEEbib}
\bibliography{strings,refs}

\end{document}